# *LADAR-Based Vehicle Tracking and Trajectory Estimation for Urban Driving*

*Daniel Morris, Paul Haley, William Zachar, Steve McLean*
**General Dynamics Robotic Systems,**
Tel: 410-876-9200, Fax: 410-876-9470
www.gdrs.com
{dmorris,phaley,wzachar,smclean}@gdrs.com

## *Abstract*

Safe mobility for unmanned ground vehicles requires reliable detection of other vehicles, along with precise estimates of their locations and trajectories. Here we describe the algorithms and system we have developed for accurate trajectory estimation of nearby vehicles using an onboard scanning LADAR. We introduce a variable-axis Ackerman steering model and compare this to an independent steering model. Then for robust tracking we propose a multi-hypothesis tracker that combines these kinematic models to leverage the strengths of each. When trajectories estimated with our techniques are input into a planner, they enable an unmanned vehicle to negotiate traffic in urban environments. Results have been evaluated running in real time on a moving vehicle with a scanning LADAR.

## *1. Introduction*

Vehicle detection, tracking and trajectory estimation are crucial components of the mobility system for unmanned ground vehicles (UGVs). There has been significant work in this area for highway driving where the lane network is well structured, see [12] for an overview. Much less work addresses the problems of urban driving with the need to negotiate with close-in maneuvering vehicles, parked vehicles and other objects. This work focuses on tracking and estimating the trajectories of nearby vehicles in urban type of clutter.

LADAR sensing is one of the most useful modalities for detecting and locating stationary objects, see [1], [3], [4] and [13]. It provides accurate day/night range estimates that can be readily accumulated to enable categorization [4]. However, application to moving objects has proven more difficult due to the registration difficulty. Detection and tracking based on global segmentation [2] along with cluster similarity [9], [11] and feature detection [8], [10], [14] have been proposed. In all these cases very simple motion models are used that do not take advantage of the nonholonomic kinematic constraints of vehicles.

In this paper we investigate how incorporating target vehicle kinematics and shape information can improve the accuracy and robustness of vehicle tracking from a UGV. To do this we introduce a *variable-axis Ackerman steering model* (VASM) and compare this with the more standard *independent steering model* (ISM). These models are adaptive and general enough for tracking most vehicle types. Then we propose a combined multi-hypothesis tracker that leverages the strengths of each of these to gain both efficiency and robustness. We show results of real-time vehicle tracking from a moving UGV.

## *2. Vehicle Modeling*

We address both kinematic and shape modeling of vehicles. Our two different kinematic models are described in Sections 2.1 and 2.2. For shape modeling we assume an adjustable rectangular outline for vehicles aligned with the vehicle orientation whose dimensions are incrementally estimated as described in Section 2.3.

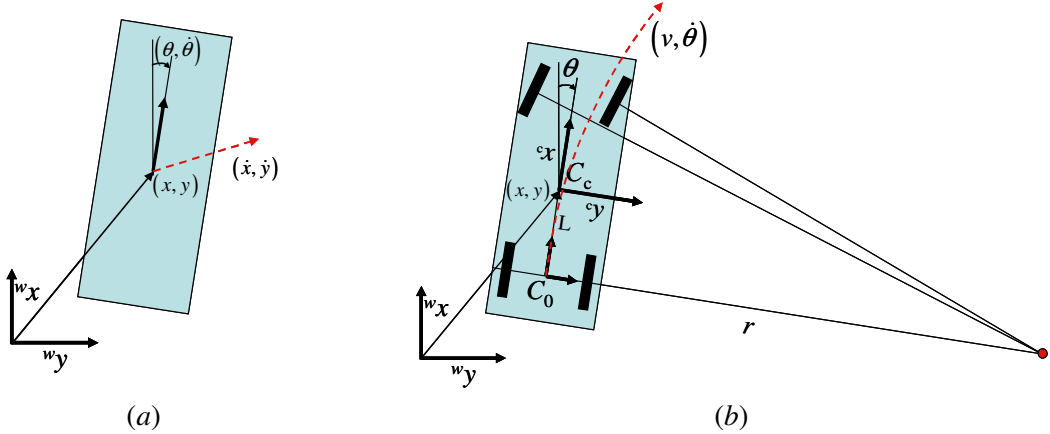

(a)  (b)

**Figure 1** Parameters for (*a*) ISM and (*b*) VASM. Vehicle velocity is independent of vehicle orientation in the ISM but is constrained to follow a particular arc in the VASM.

**2.1 Independent Steered Model (ISM)**

The ISM is so called since its velocity direction is not constrained by its orientation, see Figure 1(*a*). It is described by a 6-parameter constant velocity model with a state vector, $x(t_k)$, containing position, *x, y*, orientation, $\theta$, and their time derivatives, along with a state transition matrix, $\Phi(t_k)$:

$$x = \begin{pmatrix} x \\ \dot{x} \\ y \\ \dot{y} \\ \theta \\ \dot{\theta} \end{pmatrix}, \qquad \Phi = \begin{pmatrix} 1 & \Delta t & & & & \\ & 1 & & & & \\ & & 1 & \Delta t & & \\ & & & 1 & & \\ & & & & 1 & \Delta t \\ & & & & & 1 \end{pmatrix}, \qquad (1)$$

where for notational simplicity the time, $t_k$, is dropped. A system noise matrix can be derived by assuming an unknown continuous white noise acceleration in the interval $\Delta t$. Treating the $x, y, \theta$ components independently, the differential state transition matrix for the $(x \ \dot{x})^T$ component is $F_x = \begin{pmatrix} 0 & 1 \\ 0 & 0 \end{pmatrix}$, and so $Q_x$ can be expressed as the solution to the differential equation:

$$\dot{Q}_x = F_x Q_x + Q_x F_x^T + \begin{pmatrix} 0 & 0 \\ 0 & \alpha \end{pmatrix}, \quad (2)$$

where $\alpha$ is the variance of the white noise acceleration term. Given that $Q_x$ is symmetric, we integrate its components over the interval $\Delta t$ to obtain:

$$Q_x = \alpha \begin{pmatrix} \Delta t^3/3 & \Delta t^2/2 \\ \Delta t^2/2 & \Delta t \end{pmatrix}. \quad (3)$$

The same procedure is followed to obtain $Q_y$ and $Q_\theta$, which are identical except that $Q_\theta$ has a different white noise variance, $\beta$. Hence the full system noise is:

$$Q = \begin{pmatrix} Q_x & & \\ & Q_y & \\ & & Q_\theta \end{pmatrix}. \quad (4)$$

**2.2 Variable-axis Ackerman Steered Model (VASM)**

The VASM moves along an arc illustrated in Figure 1(b) and is specified by a six parameter state vector: $x = \begin{pmatrix} x & y & L & v & \theta & \dot{\theta} \end{pmatrix}^T$. Here *x, y*, is the center position, *L* is the location along the centerline of the rotation axis in the vehicle-centered coordinates, *v* is the arc speed, and $\theta, \dot{\theta}$ are the orientation and angular speed respectively. One of the challenges in using VASM is that curved trajectories are quite sensitive to the position of rotation axis, *L*, which for most cars is the unknown location of the rear-wheel axis. By estimating this, our model can adapt to a variety of vehicles, including skid-steering vehicles that can have a varying rotation axis. Estimating *L* also enables the model to adjust to either forward or backward vehicle orientations simplifying initialization.

While the position is in world coordinates, it is easiest to derive the state transition matrix in the local coordinate systems, $C_0$ and $C_c$, defined in Figure 1(b), and then to transform to world coordinates. We use the notation $^0x$, and $^cx$ to denote positions in these coordinate systems. First consider a point at the origin of the $C_0$ coordinate system. Its location at time $\Delta t$ is given by:

$$
\begin{aligned}
{}^0x(\Delta t) &= \int_0^{\Delta t} v\cos(\dot{\theta}t)dt = v\Delta t\,\text{sinc}(\dot{\theta}\Delta t) \\
{}^0y(\Delta t) &= \int_0^{\Delta t} v\sin(\dot{\theta}t)dt = v\dot{\theta}\Delta t^2\,\text{sinc}^2(\dot{\theta}\Delta t/2)/2
\end{aligned}
\tag{5}
$$

where we are using a constant speed, $v$, and constant angular speed, $\dot{\theta}$, assumption. If instead we work in the vehicle center coordinate system and calculate the center location at time $\Delta t$ we get:

$$
\begin{aligned}
{}^cx(\Delta t) &= {}^0x(\Delta t) + L - L\cos(\dot{\theta}\Delta t) = v\Delta t\,\text{sinc}(\dot{\theta}\Delta t) + 2L\sin^2(\dot{\theta}\Delta t/2) \\
{}^cy(\Delta t) &= {}^0y(\Delta t) - L\sin(\dot{\theta}\Delta t) = v\dot{\theta}\Delta t^2\,\text{sinc}^2(\dot{\theta}\Delta t/2)/2 - L\sin(\dot{\theta}\Delta t)
\end{aligned}
\tag{6}
$$

For simplicity we continue to let $\theta$ be the orientation of the vehicle, and $v$ be the speed of the rear axle (or more specifically, the point on the centerline whose velocity is parallel to the orientation, $\theta$). The velocity at the vehicle center will have an orientation offset $\Delta\theta_c$ and speed $v_c$ given by:

$$
\Delta\theta_c = \arctan 2(-L\dot{\theta}, v), \qquad v_c = v/\cos(\Delta\theta_c).
\tag{7}
$$

Now the local coordinate update step in Eq. (6) can be transformed to world coordinates as an update from time $t$ to $t + \Delta t$:

$$
\begin{pmatrix} x_{t+\Delta t} \\ y_{t+\Delta t} \end{pmatrix} = \begin{pmatrix} x_t \\ y_t \end{pmatrix} + R(\theta_t)\begin{pmatrix} {}^cx(\Delta t) \\ {}^cy(\Delta t) \end{pmatrix}, \text{ where } R(\theta_t) = \begin{pmatrix} \cos(\theta_t) & -\sin(\theta_t) \\ \sin(\theta_t) & \cos(\theta_t) \end{pmatrix}
\tag{8}
$$

From this we obtain the discrete state transition matrix:

$$
\boldsymbol{\Phi}(\Delta t) = \boldsymbol{I} + \boldsymbol{T}(\theta_t)\begin{pmatrix}
0 & 0 & \partial^c x/\partial L & \partial^c x/\partial v & -{}^cy & \partial^c x/\partial\dot{\theta} \\
0 & 0 & \partial^c y/\partial L & \partial^c y/\partial v & {}^cx & \partial^c y/\partial\dot{\theta} \\
0 & 0 & 0 & 0 & 0 & 0 \\
0 & 0 & 0 & 0 & 0 & 0 \\
0 & 0 & 0 & 0 & 0 & \Delta t \\
0 & 0 & 0 & 0 & 0 & 0
\end{pmatrix}
\tag{9}
$$

Where the transformation to world coordinates is:

$$\boldsymbol{T}(\theta_t) = \begin{pmatrix} R(\theta_t) & 0 \\ 0 & \boldsymbol{I}_4 \end{pmatrix} \tag{10}$$

and the partials are:

$$\begin{aligned}
\frac{\partial {}^c x}{\partial L} &= 2\sin^2(\dot\theta \Delta t/2), & \frac{\partial {}^c y}{\partial L} &= -\sin(\dot\theta \Delta t) \\
\frac{\partial {}^c x}{\partial v} &= \Delta t\,\mathrm{sinc}(\dot\theta \Delta t), & \frac{\partial {}^c y}{\partial v} &= \dot\theta \Delta t^2 \mathrm{sinc}^2(\dot\theta \Delta t/2)/2 \\
\frac{\partial {}^c x}{\partial \dot\theta} &= v\Delta t^2 \mathrm{sinc}'(\dot\theta \Delta t) + L\Delta t\sin(\dot\theta \Delta t) \\
\frac{\partial {}^c y}{\partial \dot\theta} &= v\Delta t^2 \left(\mathrm{sinc}(\dot\theta \Delta t/2)\cos(\dot\theta \Delta t/2) - \mathrm{sinc}^2(\dot\theta \Delta t/2)/2\right) - L\Delta t\cos(\dot\theta \Delta t)
\end{aligned} \tag{11}$$

The system noise matrix, $\boldsymbol{Q}$, is derived in an analogous way to that for ISM. By integrating in a local curvilinear coordinate system that follows the arc of the vehicle three independent components: $\boldsymbol{Q}_{xv}$, $\boldsymbol{Q}_{y\theta\dot\theta}$ and $\boldsymbol{Q}_L$ are isolated. $\boldsymbol{Q}_L$ is a scalar that is ideally zero for a fixed rotation axis vehicle, but nevertheless we set to a small constant, $\varepsilon_L$, to capture imperfections in our steering model as well to model vehicles with varying rotation axis. $\boldsymbol{Q}_{xv}$ is the integration of a white noise acceleration along the motion arc and thus is identical to $\boldsymbol{Q}_x$ in (3), but in this case acting on $v$ and $x$. $\boldsymbol{Q}_{y\theta\dot\theta}$ integrates a white noise angular acceleration which gives a second order noise term perpendicular to the arc, namely in direction ${}^c y$. Performing this integration gives a term:

$$\boldsymbol{Q}_{y\theta\dot\theta} = \gamma \begin{pmatrix} v^2 \Delta t^5/20 & |v|\Delta t^4/8 & |v|\Delta t^3/6 \\ |v|\Delta t^4/8 & \Delta t^3/3 & \Delta t^2/2 \\ |v|\Delta t^3/6 & \Delta t^2/2 & \Delta t \end{pmatrix}. \tag{12}$$

Putting these terms together and shifting rows to align with the state vector we get:

$$Q = T(\theta_t) \begin{pmatrix} \alpha\Delta t^3/3 & 0 & 0 & \alpha\Delta t^2/2 & 0 & 0 \\ 0 & \gamma v^2 \Delta t^5/20 & 0 & 0 & \gamma|v|\Delta t^4/8 & \gamma|v|\Delta t^3/6 \\ 0 & 0 & \varepsilon_L & 0 & 0 & 0 \\ \alpha\Delta t^2/2 & 0 & 0 & \alpha\Delta t & 0 & 0 \\ 0 & \gamma|v|\Delta t^4/8 & 0 & 0 & \gamma\Delta t^3/6 & \gamma\Delta t^2/2 \\ 0 & \gamma|v|\Delta t^3/6 & 0 & 0 & \gamma\Delta t^2/2 & \gamma\Delta t \end{pmatrix} T(\theta_t)^T \quad (13)$$

where we have rotated into world coordinates. We note that this system noise matrix is calculated for the center rear axle location, but we expect this to be a very good approximation to that at the vehicle center.

**2.3 Vehicle Shape Model**

Our vehicle model is a 2D rectangle centered at the vehicle location and oriented with the main edge of the vehicle. After robust fitting at each time step (Section 2.4) we obtain measurements of the width and/or length. Each length measurement is added to a normalized histogram of lengths (where the normalization determines the half-life). To eliminate large outliers, the largest 5% of measurements are excluded. But the main outliers to exclude are from measurements taken during partial occlusions which can generate large peaks in the histogram. These peaks will always be at shorter lengths than the true length, and so they can be excluded by choosing the peak with greatest length. The same procedure is used to estimate width. In this way we maintain a robust and adaptive shape estimate.

**2.4 Robust Data Fitting**

At any time at most two sides of a vehicle are visible. This means there are two fitting scenarios: either a single edge or a perpendicular pair of edges as illustrated in Figure 2. In each case an estimate on the vehicle position and orientation is obtained giving a measurement vector and matrix:

$$z = \begin{pmatrix} x \\ y \\ \theta \end{pmatrix}, \text{ and } H = \begin{pmatrix} 1 & 0 & 0 & 0 & 0 & 0 \\ 0 & 1 & 0 & 0 & 0 & 0 \\ 0 & 0 & 0 & 0 & 1 & 0 \end{pmatrix}. \qquad (14)$$

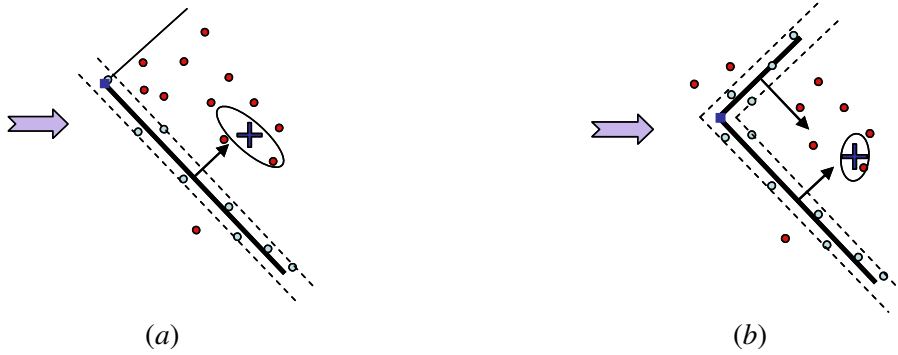

(a)            (b)

**Figure 2** Robust data fitting of an edge (thick dark line) (*a*), or a "L-shape" (*b*). The inlier LADAR points (shown in green) are used to improve the fit, which in each case defines a corner position and orientation. Using this corner, the vehicle center is estimated along with a covariance.

We use a RANSAC-based technique to robustly fit edges and corners to LADAR points on the vehicle. Two points are needed for a line hypothesis, and three points for a perpendicular corner. The corner fitting can be made more efficient and robust by taking into account visibility constraints illustrated in Figure 3:

- The corner must be visible from the exterior, and
- Neither edge can occlude the other.

These are encoded in the following visibility condition. Let $\theta_c$ be the viewing direction, and $\theta_a$ and $\theta_b$ be the orientations of the two perpendicular edges from the corner. Both edges are visible if and only if:

$$|\theta_c - \theta_a| < 90° - s \quad \text{and} \quad |\theta_c - \theta_b| < 90° - s, \qquad (15)$$

where $s$ is the minimum slant angle at which an edge is visible, in our case 10 degrees.

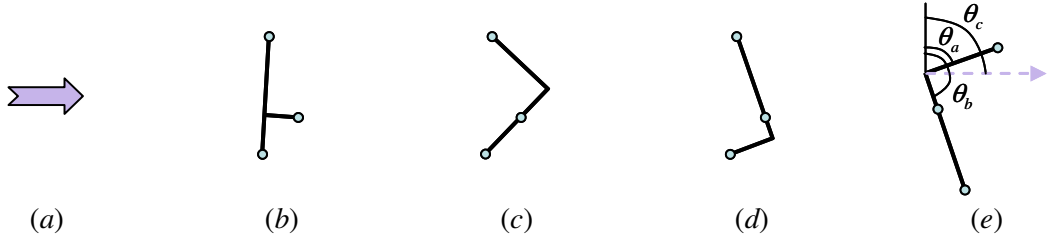

(a)          (b)          (c)          (d)          (e)

**Figure 3** Three non-colinear points define three distinct perpendicular corners, see (b), (c) and (d). In this case physical considerations exclude all three, given the viewing direction shown in (a). (e) shows a valid corner passing the conditions in Eq. (15).

Once inliers to a corner are found, the corner parameters are estimated by minimizing the squared perpendicular distance of the points to the edges:

$$J(x_c, y_c, \phi) = \sum_{i \in inliers(edge1)} \left(\cos(\phi)(x_i - x_c) + \sin(\phi)(y_i - y_c)\right)^2 / 2 + \sum_{j \in inliers(edge2)} \left(\cos(\phi)(y_j - y_c) - \sin(\phi)(x_j - x_c)\right)^2 / 2 \quad (16)$$

Here $(x_c, y_c, \phi)$ encode the corner location and orientation. The Hessian can be approximated as the 3x3 matrix: $\nabla^2 J \approx \nabla J^T \nabla J$, and we found that a single Gauss-Newton step was sufficient to optimize the result from RANSAC.

The final step in fitting is to convert the corner fit into a measurement vector, $z$, and covariance, $R$. Using the current width and length estimates of the vehicle and the fitted orientation, $\varphi$, let $(x_d, y_d)$ be the offset from the corner to the vehicle center, and $\tau$ be the rotation in multiples of 90° that aligns the corner edge with the appropriate vehicle edge. Then using the Gauss-Newton optimized corner parameters, we obtain:

$$z = \begin{pmatrix} x_c + x_d \\ y_c + y_d \\ \varphi + \tau \end{pmatrix}. \quad (17)$$

Now a covariance estimate for the corner location can be obtained as the inverse of the Hessian. However, we need the covariance of the vehicle center which may have significant correlation effects due to being offset from the corner. But this is easy to find; we simply evaluate the Hessian using the vehicle center coordinates rather than the corner coordinates:

$$R = \sigma^2 \left( \nabla J(z)^T \nabla J(z) \right)^{-1}, \quad (18)$$

where $\sigma^2$ is the variance of the LADAR returns.

**2.5 Multi-Hypothesis Tracking**

All of the components for building a Kalman Filter-based tracker have been described. These components are brought together with the following standard equations:

$$\begin{aligned}
x_t^{(-)} &= \Phi_t x_{t-1}^{(+)} \\
P_t^{(-)} &= \Phi_t P_{t-1}^{(+)} \Phi_t^T + Q_{t-1} \\
K_t &= P_t^{(-)} H_t^T \left( H_t P_t^{(-)} H_t^T + R_t \right)^{-1} \\
x_t^{(+)} &= x_t^{(-)} + K_t \left( z_t - H_t x_t^{(-)} \right) \\
P_t^{(+)} &= P_t^{(-)} - K_t H_t P_t^{(-)}
\end{aligned} \quad (19)$$

A key step that remains is choice of model. The VASM better captures constraints on vehicle motion. It is has less bias and more accurate trajectory prediction, especially in curves as illustrated in Figure 4. However, it has two main drawbacks. (1) Given the detection of a corner of a stationary vehicle there is a 90 degree ambiguity in orientation. This means vehicles would need to be initialized with two states 90 degrees apart. (2) The flip side of having more constrained motion is that if spurious data cause the model to become misaligned with the true state, it is more difficult for it to recover than ISM.

In urban environments most objects detected will be stationary clutter or stationary vehicles. To avoid needing two models for each of these, we initialize all objects with a single ISM. Objects are tracked and any that are determined to be movers are subsequently initialized with an additional VASM. In this way multiple hypothesis tracking is only applied to movers. Having both an ISM and VASM and switching between them based on fitting score gives robustness to each of their weaknesses. If one track fails to fit for a few frames it is reinitialized with the other.

There are a few possible errors in fitting including when the edge or corner is fit to internal structure of the vehicle or to clutter points grouped in with the vehicle. Cases like these can be handled by adding constraints to the model fitting. However, adding constraints makes the model less flexible and potentially creates additional failure modes. To get the benefits without the harm of constraints, an additional VASM track hypothesis is added for each constraint. A track manager tracks each hypothesis independently and at any given time selects the best model as the current target state. In our implementation we maintain up to four tracks per mover and use the additional tracks to vary the fraction of points that must lie inside the vehicle boundary, and also vary the scale factor on the system noise which implicitly adjusts how much we trust the model prediction versus the measurements.

## *3. Results*

The tracker has been tested on a large number of scenarios including moving sensor platform, multiple target vehicles and high clutter examples. Figure 4 illustrates how the VASM model has reduced bias and better trajectory estimation than ISM for turning vehicles. However, using ISM for tracking stationary objects improves efficiency since only one track per stationary object is needed, see Figure 5. Also ISM is less likely to suffer track loss with noisy measurements and so by combining ISM and VASM we are able to robustly track movers even for very noisy scenarios, see Figure 6.

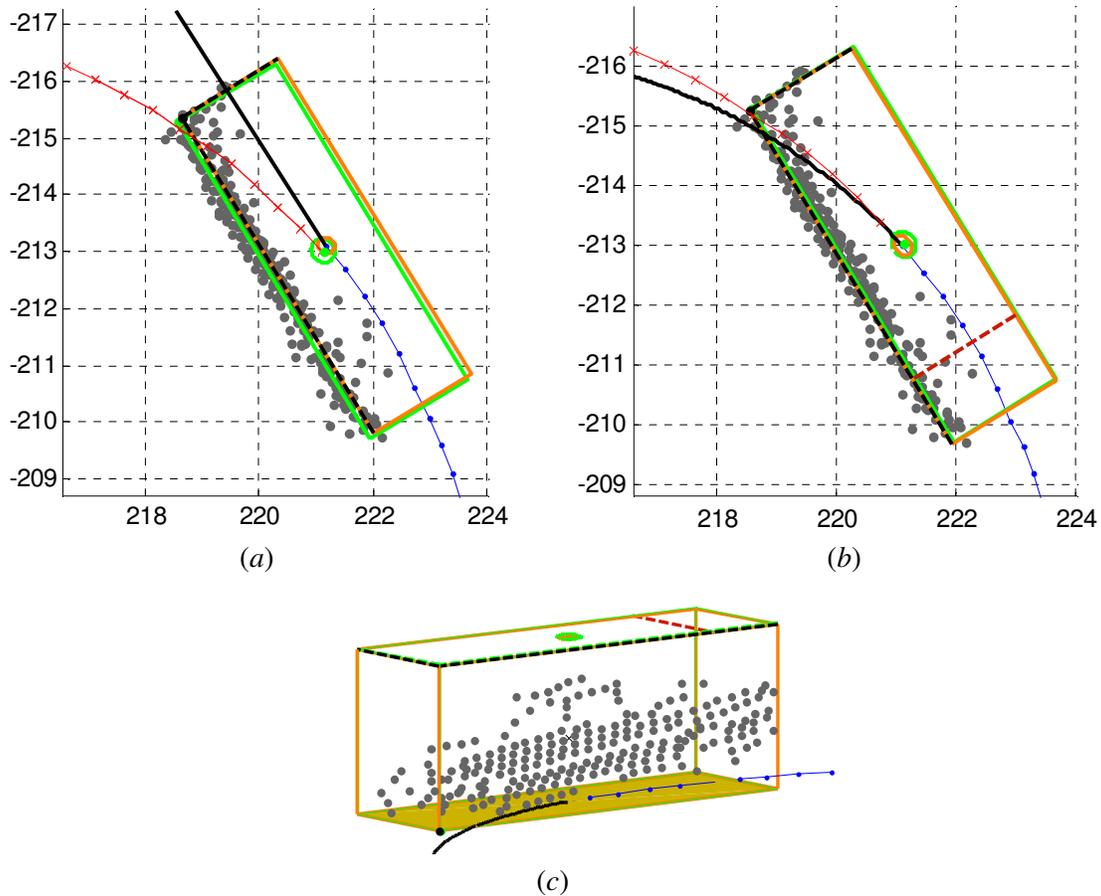

(*a*)   (*b*)

(*c*)

**Figure 4** Comparison of vehicle tracking with ISM in (*a*), and VASM in (*b*) and (*c*), where the latter shows a 3D plot. The gray dots are LADAR hits, the green rectangle and ellipse represent the measurement, *z*, along with covariance in position. The black dashed lines show which edges are visible and fit. The orange rectangle and ellipse show the updated state vector plus position covariance. In (*b*) and (*c*) the dashed red line is the estimated location of the rear axle (*L*). The blue line leading up to the current position is the recent vehicle trajectory and the black line leading from this is the predicted trajectory which can be compared to the true trajectory shown in red.

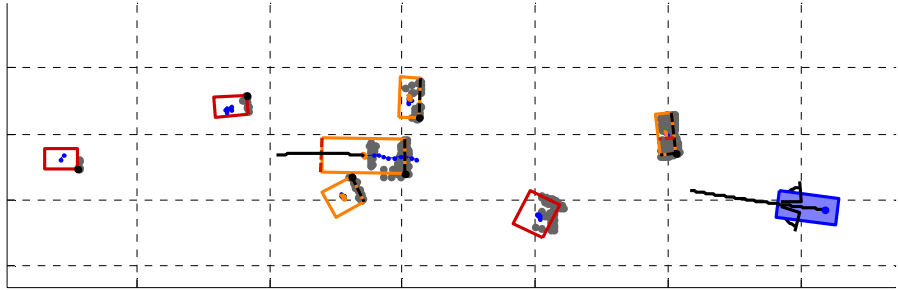

**Figure 5** A scenario in which a target vehicle is being followed by a sensor vehicle (blue). The target vehicle is maneuvering through clutter objects that are being tracked with ISM. The target vehicle itself is being tracked with both ISM and VASM. Note in this case, where only noisy rear-end measurements are obtained, the rotation axis is incorrectly estimated near the front of the vehicle, although when the target vehicle turns sharply the axis shifts to the correct location.

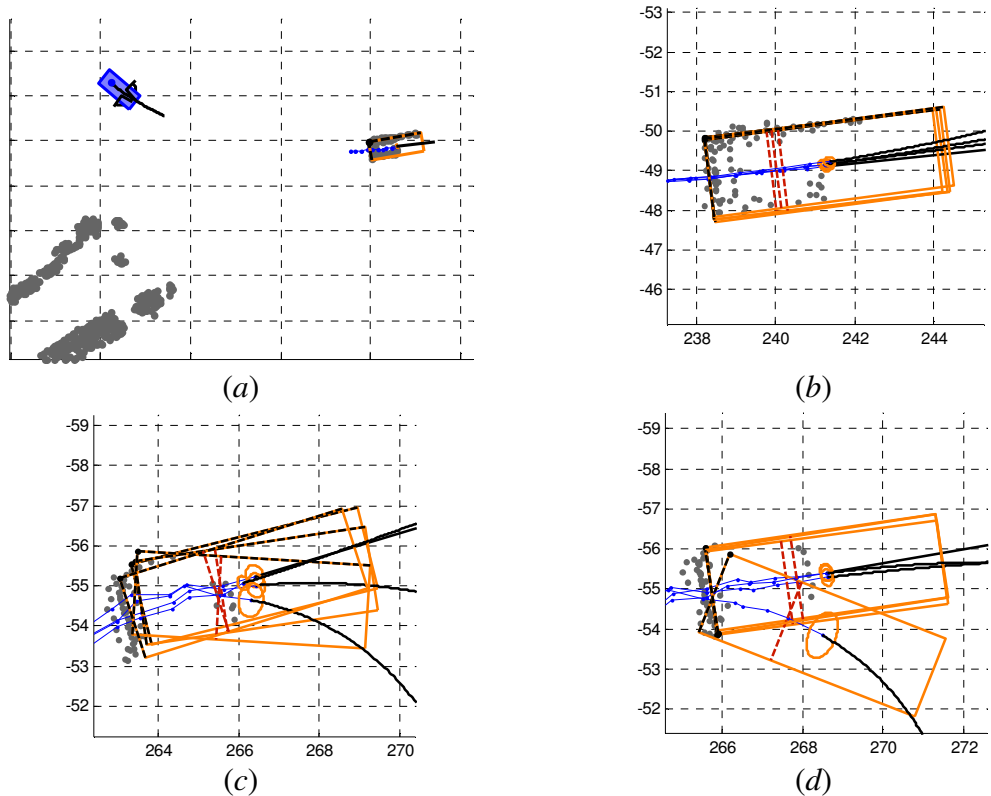

**Figure 6** Scenario in which a target vehicle is being followed and tracked using the multi-hypothesis tracker. (*a*) A view of the sensor vehicle (blue) following the target vehicle. (*b*) A 4-hypothesis tracker with 1 ISM and 3 VASMs. (*c*) An ambiguous fitting example leads to one of the VASMs diverging. If it diverges sufficiently, as shown in (*d*), it fails to fit, is lost and needs to be reinitialized using one of the good hypotheses.

## *4. Conclusion*

We have developed a vehicle tracking and trajectory estimation approach that involves adaptively estimating vehicle shape along with the kinematic parameters. We introduced VASM, a kinematic model that can capture Ackerman steering when the rear axis is unknown as well as model skid-steering. We compared it to ISM and identified the strengths of each. Then we proposed a multi-hypothesis tracker that combines both of these to achieve both high efficiency and improved robustness.

In the future we plan a quantitative comparison of these models using ground truth. We also plan to extend them to model vehicles with trailers.

## *Acknowledgements*